\documentclass{article}
\usepackage[utf8]{inputenc}
\usepackage{graphicx}
\usepackage{authblk}
\usepackage[scale=0.8]{geometry}
\usepackage{cite}

\title{Spiking Associative Memory For Spatio-Temporal Patterns}
\author{Simon Davidson, Stephen B. Furber, Oliver Rhodes}

\affil{The University of Manchester, UK}

\date{Correspondence: \{simon.davidson, oliver.rhodes\}@manchester.ac.uk\newline\newline June 2020}

\begin{document}

\maketitle

\begin{abstract}
Recent successes in the application of neural networks have been in Artificial Neural Networks (ANNs) with deep structure and trained with back-propagation, inspired by neuroscience but very much an engineering abstraction. Traditional approaches more firmly rooted in the biology - with information flow encoded as action potentials and learning using biologically-inspired rules such as Spike Time-Dependent Plasticity (STDP) - have failed to deliver results comparable in capability and performance to an ANN running on a GPU. The lack of reliable principles for representation, computation and learning in spiking neural networks is arguably an essential missing piece, without which the design of complex and structured neural systems becomes a near impossible task. How are these shortcomings to be addressed?

We make the assumption that spiking neural networks can only be efficient if some form of coding using the precise time of individual spikes is employed, rather than the rate coding that typifies most research in this area. Furthermore we suggest that fast reliable associative memory capable of working in noisy environments is a key building block of larger neural systems. Together these constraints motivate the two major themes explored in this paper. The first is the development a simple stochastic learning rule called \emph{cyclic STDP} that can extract patterns encoded in the precise spiking times of a group of neurons, subject both to jitter in the timing of individual spikes and to the presence of background noise in the form of unwanted spikes embedded in the signal. We show that a population of neurons endowed with this learning rule can act as an effective short-term associative memory, storing and reliably recalling a large set of pattern associations over an extended period of time. We require only that associated information is encoded in a periodic spatio-temporal format called \emph{cyclic N-of-M encoding}, a sparse fixed-weight representation framework whose design philosophy builds on the Correlation Matrix Memory (CMM), extended into the temporal domain.

The second major theme examines the challenges associated with training a neuron to produce a spike at a precise time and for the fidelity of spike recall time to be maintained as further learning occurs. Such considerations are of immediate concern if we wish to formulate rules of neural computation that are sensitive to the precise timing of individual spikes. The strong constraint of working with precisely-timed spikes (so called \emph{temporal coding}) is mandated by the learning rule but is also consistent with the believe in the necessity of such an encoding scheme to render a spiking neural network a competitive solution for flexible intelligent systems in continuous learning environments.

The encoding and learning rules are demonstrated in the design of a single-layer associative memory (an input layer consisting of 3,200 spiking neurons fully-connected to a similar sized population of memory neurons), which we simulate and characterise. Design considerations and clarification of the role of parameters under the control of the designer are explored.

 \textbf{Keywords:} cyclic\ STDP\ , precise\ spike\ timing\ , cyclic\ N-of-M\ encoding\ , Poisson\ filter, locked\ binary\ weights
\end{abstract}

\section{Introduction}
The last decade has seen an dramatic increase in the capabilities and performance of so-called {Deep Artificial Neural Networks} (Deep ANNs) that are inspired by but highly abstracted from the biological neural networks of real brains. The ability to train very many layers of neurons successfully has been the key enabler, built around changes to the activation function (which defines the continuous-valued response produced by each neuron) and new techniques used to normalise the collective outputs of entire populations of neurons so that their joint output is as informative as possible. Research into more biologically-faithful networks, in which neurons produce action potentials (or \emph{spikes}) and are connected only sparsely to the groups of neurons that they target, is receiving less and less attention. It might be argued that these biological aspects of network design are superfluous and inefficient hang-overs of evolution to be discarded in the same way that the design of modern aircraft does not incorporate flapping wings.

One route to overcome this pessimism is by demonstrating the converse - to develop a toolkit of neural network components based around spiking neurons that form a framework for the principled design of superior solutions to commercially relevant problems. If the development of these components sheds any light on the operating principles of the biological networks that inspired them, we should also grasp this opportunity with alacrity. This paper represents small steps towards this goal of a set of design principles for large neural systems based on spiking neurons and biologically-plausible learning.

Let us outline some basic assumptions not all of which we will attempt to prove in this paper, although even those that are not proven should at least appear reasonable. First is the assertion that the use of precise spike times should be fundamental requirement for pattern encoding, storage and computation, in contrast to other accepted encoding schemes such as \emph{rate-based} encoding (where the exact time of generation of spikes by a neuron conveys no information, only their rate). It should be uncontentious to justify this purely on the grounds of the energetic efficiency of communication compared to that of rate-based encoding, but one could also appeal to the nature of the computation performed by the artificial neurons themselves, which fire when sufficient input activity has been summed in a given window of time. Metrics based on spike co-incidence are more meaningful when the timing of the spikes themselves has significance and is being actively managed.

Our second assumption is that the memory of a flexible neural agent must cope with a stream of incoming data items. The lifetime of each may be short or long and their place in the world model of the agent may not be clear when they are first acquired. Consolidating incoming data into memory is therefore a more complex task than simply accepting everything equally, instead consisting of the gradual absorption of information over multiple time scales, subjecting it to filtering and re-coding throughout this process. To manage this challenging task we assume the existence of multiple levels of memory forming a hierarchy, from reliable but low-capacity short-term memory through to optimised, high-capacity long-term memory. For reasons of practicality and energy efficiency state for each of these memory structures will be distributed, co-existing in each synapse rather than as disjoint cortical regions.

The need to balance the reliability of absorption against the risk of destroying existing stored information is known as the \emph{stability-plasticity dilemma} and is a major challenge for all types of neural networks including the successful deep networks (which typically allow no further learning once initial training is complete). In this paper we are concerned only with the first level of this memory hierarchy: short-term memory (or STM).  We define its role as capturing a list of associations between \emph{key-data} pairs and recalling each \emph{data} item when its associated \emph{key} is presented. STM is the building block for rapid but transient associations as well as the gateway to more permanent storage and recall. The requirements and design for deeper levels of the memory hierarchy and the processes governing assimilation are left to future work. 

Our third assumption is that the activity of each network must cope with the presence of noise, in the form of spurious background activity (unwanted spikes) as well as \emph{jitter} (time shifting of individual spikes subject to zero mean Gaussian noise) of signal spikes. These processes are long studied \emph{in vivo} and may be the result of nothing more than the chatter of molecular mechanisms that are ever-present in real cells. These unwanted spikes are assumed to be Poisson distributed: independent and characterised only by their mean spiking rate.

Our final assumption is that learning rules should be biologically plausible, which in practise means that they should use only a combination of locally available information and globally broadcast signals. This rules out learning that involved implausible calculations such as the inversion of large matrices. This assumption has practical benefit since learning based on these principles will be easier to implement in large scale systems.

While deep networks have used variants of error back-propagation for several decades, few successful learning rules have emerged for spiking networks, partly because of the inability to perform differentiation on sequences of spikes. Neuroscience has furnished the machine learning community with some learning rules over the years, with Spike Time-Dependent Plasticity (STDP) being the most studied since it was isolated \emph{in vivo} during the 1990's \cite{markram96,bi98synaptic}. This rule combines traces of activity in both the pre-synaptic and post-synaptic spike trains to determine changes to the weight value in each synapse. However, as a mechanism for long-term memory it is far from ideal: stored information is typically erased too quickly by both new activity and background noise inherent in large networks to be of use in practical systems. Yet STDP (in some form) has been demonstrated in isolated brain tissue - how do we reconcile this apparently conflicting evidence?

\subsection{Dissecting STDP}
Although there are many variants and modes of STDP that have been proposed \cite{markram12}, the basic premise of STDP at a synapse is that changes to its efficacy are made based on the detection of causal and anti-causal relationships between its pre-synaptic and post-synaptic spike trains. Specifically, if a pre-synaptic spike arrives and some short time later a post-synaptic spike is generated this is interpreted as a causal relationship and the strength of the synaptic weight is increased (the synapse is \emph{potentiated}). If the order of arrival of the spikes is reversed, this is interpreted as an anti-causal relationship and the synaptic strength is reduce (the synapse is \emph{depressed}).

The erroneous interpretation of STDP is that all spike activity (both pre- and post-) contributes to changes in a synaptic weight. This occurs even in the presence of spurious background spikes. As a result, a carefully crafted synaptic weight value is subject to drift over time due to new activity including noise, forgetting the information it was required to store. And yet the early \emph{in vitro} experiments are clear that single pairs of spikes do not cause changes in synaptic weight on their own \cite{morrison08}. Instead, spike streams consisting of tens or hundreds of pairs of spikes were required to elicit synaptic change. This observation alone suggests a subtle but important re-interpretation of how STDP functions, which makes stable storage of multiple items possible but introduces a significant restriction on how information should be coded in neocortex.

\subsection{Proposed Learning Rule and Encoding - Cyclic STDP and Cyclic N-of-M Patterns}
Consider the world from the point of view of a single synapse that receives a stream of spikes on its input and is able to discern that its parent neuron is also producing a stream of spikes through some signal that back-propagates from the neuron as it fires.
Both the input and output spike trains are subject to noise (due to jitter and the presence of spikes unrelated to the intended signal). It is desirable that the synapse responds to signal from an information-carrying pattern, but ignores noise. What information can it use to make this determination?

Clearly if the patterns on both the input and output were to be repeated, the signal spikes would occur again with the same phase relationships (inputs to outputs) they did on the first presentation (perhaps subject to some jitter). But the background noise spikes would have a much lower probability of occurring again at the same time as on the first repeat of the patterns. with each additional repeat of the input-output pattern-pair, the task of identifying the true signal spikes becomes easier. This is the strategy adopted here. Patterns are constrained to be spatio-temporal in nature, of a fixed cycle length and with each spike occurring at a fixed time within the cycle. The STDP-based learning process is now recast as one of extracting signal spikes from a background of Poisson-distributed noise spikes.

The learning prescription demands that the input and output patterns be presented multiple time in succession, after which they can be discarded. The learnt association will then be stable until it is explicitly erased - any presentation of a previously seen input pattern will trigger each neuron to produce its associated output spike at the time it was taught. Each synapse has hidden state that allows it to gather evidence that the spike trains in its pre- and post-synaptic neurons are either \emph{causally} or \emph{anti-causally} related. When this evidence reaches a defined threshold (after multiple repeats of the pattern pairs) this triggers modification to the synaptic weight for this connection. \textbf{If this threshold is not reached, no such change occurs and the synapse is protected from erosion of its weight}.

The accumulation of evidence in each synapse is similar to (but distinct from) what is called a \emph{synaptic trace} in the literature. Typically, a trace is a counter whose value is incremented by a fixed amount with the arrival of each pre-synaptic spike, but decays towards zero over time. Hence the value of the trace carries a compact summary of the recent history of the input spike train to that synapse. When a post-synaptic spike occurs the current value of the trace is used to determine the magnitude of the synaptic weight change. In the proposed learning scheme the subtle but crucial difference is that the information of interest is not in any one spike train, but rather in the occurrence of pre-post or post-pre \emph{spike pairs}, signalling possible causal or anti-causal relationships (respectively) between the two spike trains.

The next section described the theory in detail, beginning with the nature and properties of the patterns being associated, the synaptic state machines that determine that a change is necessary and the rules determining the specific change required. Section \ref{Experiments} describes experiments that demonstrate the action of the learning process. Section \ref{Discussion} summarises the key points of the research and the possible applicability of the main ideas to explanations of phenomena observed in neocortex. The section concludes with suggestions for future research directions building on the new learning rule.

\section{Methodology \& Theory} \label{Theory}
\subsection{Cyclic N-of-M Patterns} \label{cyclicNofMpatterns}
Patterns of activity across populations of neurons can be viewed abstractly as codewords. A cyclic N-of-M codeword, \emph{x}, is a periodic spatiotemporal pattern of period T ms across M binary channels, $v_i$, so \emph{x} = [$v_0, v_1, v_2,..., v_{M-1}$]. We discretise the temporal dimension so that the information content of each channel \emph{$v_i$} is itself represented using a vector of discrete time bins, each of width $T_{bin}$. Thus there are \(B=\frac{T}{T_{bin}}\) time bins in every period of each vector \emph{v} and in each bin a given channel can be either set or clear. A code weight of N means that within a given codeword exactly N of the binary bits are set during each period, i.e. N =  $\Vert \emph{x} \Vert _1$. Typically N $\ll$ M and the codewords are described as \emph{sparse}.

When translated into the context of spiking neural networks, the output of a group of M neurons is an N-spike spatiotemporal pattern repeating with period T. We will refer to patterns in this neural context from now on. The N spikes are distributed independently throughout the period, so that to an external observer seeing only a single iteration of the pattern the spikes could be the output of M independent Poisson processes. However, over multiple iterations a perfect examplar of such a pattern repeats exactly, so that it is clear that the timing of the individual spikes is deterministic.

\begin{figure}
    \centering
    \includegraphics[width=0.7\textwidth]{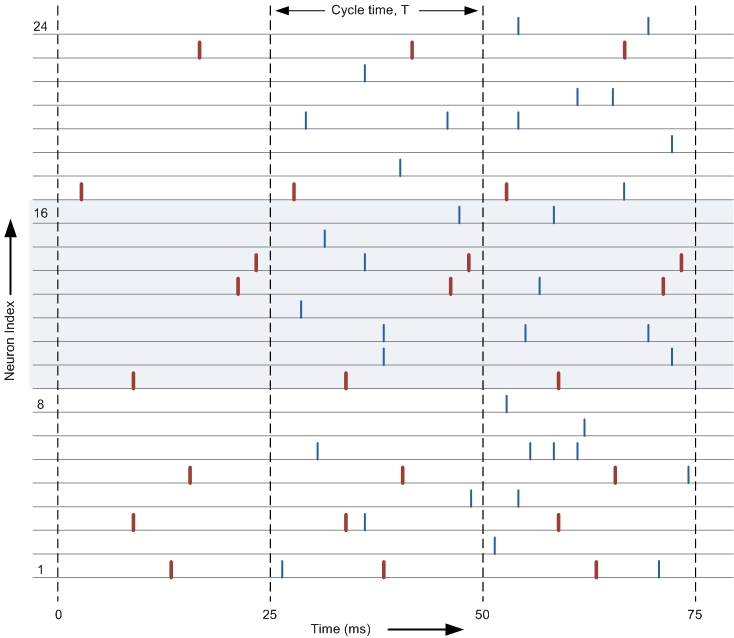}
    \caption{Graph showing three cycles of a cyclic N-of-M pattern. cycle 1 contains only signal spikes (red), while cycles 2 and 3 also include spurious background noise spikes (blue). Note that the signal spikes occur consistently in each cycle whereas the noise spikes do not.}
    \label{fig:cyclicNofMpatterns}
\end{figure}

A coarse measure of the information content, $I_{cyclic}$,  of a cyclic N-of-M codeword (in bits) has two components: (1) spatial information due to the choice of N neurons from the M available; (2) for each selected neuron, temporal information required to specify the time bin in which the spike falls:
\begin{equation}
I_{cyclic} = log_2 \left(  \frac{M!}{(M-N)! N!} \right) + N log_2 \left(  \frac{T}{T_{bin}}\right)  \label{eq:infoDynamic}
\end{equation}

The choice of a fixed weight code (with precisely N neurons firing in each pattern) has a number of advantages, not least of which is the potential for robustness if any given neuron ceases to function. It also delivers a constant quantity of 'signal' to target neurons making the task of learning to respond to the input pattern easier. This is explored in section \ref{weightChanges}. The proposed learning rule extends the concept of the Correlation Matrix Memory (CMM) \cite{will69, furb04} into the temporal domain \cite{furb07}. The CMM is an associative memory that can associate an input and output pattern rapidly and reliably provided that they conform to a set of constraints (sparse fixed weight binary codes of fixed length). The synaptic weights themselves are binary leading to very a sparse, compact connectivity matrix. The introduction of the temporal aspect to the codewords is done by assigning each output spike a time value within a fixed window, the pattern cycle time. 

The use of sparse patterns, where N $\ll$ M, leads to a low recruitment of synapses. Let $f$ = N / M represent the probability that a neuron is activated in any pattern. Then the expected probability that a given synapse will detect causal or anti-causal activity  between its input and output spike trains is only $f^2$. For 3\% activity of neurons in a pattern this represents a probability of only 0.09\% that a given synapse will be involved in any given learned association.

\subsection{Identification of Causal and Anti-casual Relationships} \label{synapticStateMachines}
The learning rule has two distinct components corresponding to the \emph{when} and the \emph{how} of learning: \emph{when} should the synapse change its connection strength and \emph{how} should it be modified? These questions can be addressed separately. We first consider the \emph{when} question.

\begin{figure}
    \centering
    \includegraphics[width=1.0\textwidth]{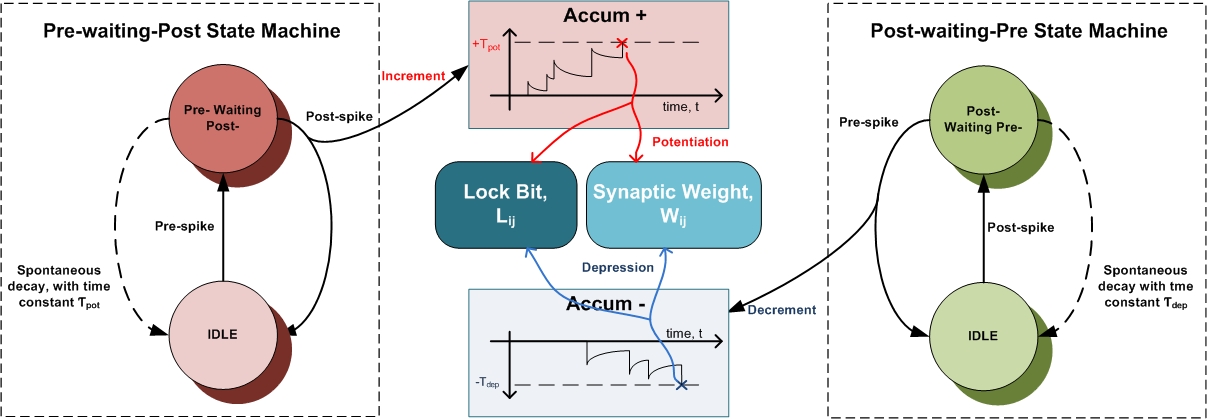}
    \caption{Schematic of the \emph{cyclic stdp state machine} in each synapse, which gathers evidence on causal and anti-causal activity in its pre- and post-synaptic spike streams.}
    \label{fig:synapticStateMachines}
\end{figure}

As described in the previous section, each synapse has a state machine whose function is to gather evidence from the pre- and post-synaptic spike trains to detect a \emph{causal} and \emph{anti-causal} relationship between them for the current input and output patterns. Figure \ref{fig:synapticStateMachines} is a schematic of this state machine built around two pieces of synaptic state: the familiar connection strength (or \emph{weight}), $W_{ij}$, and a novel state bit called the \emph{lock} bit, $L_{ij}$, whose function will be made clear in the next subsection.

On the left of the figure is a single bit state machine called \emph{Pre-waiting-post} whose task is to detect evidence of possible causal link between the pre- and post- spike trains. It remains in state \textbf{IDLE} until a pre-synaptic spike arrives, at which point it enters state \textbf{Pre-waiting-post}. If a post spike occurs while it remains in this heightened state this supports the hypothesis that there is a causal link from pre- to post- spikes. The state drops back to \textbf{IDLE} but the event is noted by incrementing a counter called the \emph{Accumulator} (denoted Accum+ in the figure). While waiting for the arriving of the post spike, the state will eventually decay spontaneously to \textbf{IDLE} without triggering an increment to the accumulator and has the effect of favouring spike pairs that have a small inter-spike interval when gathering evidence of a causal relationship. The decay process has time constant $\tau_{pot}$.

If a sufficient number of pre-post- spike pairs have been detected the accumulator reaches some threshold $T_{pot}$, triggering a potentiation event. In that case the synaptic weight and lock bit are modified (according to rules defined in the next subsection) and the accumulator is zeroed. 

The value held in the accumulator itself will decay over time, so that the spike pairs must occur in a finite window of time to trigger potentiation. Overall, a potentiation event is only possible when multiple spikes pairs with the pre-spike closely followed by the post-spike have occurred in a window of time (typically less than one second in our model).

On the right of the figure is an identically-structured state machine called \emph{Post-waiting-pre} whose function mirrors that of the first. Its task is to detect evidence of an \emph{anti-causal} relationship between the pre- and post-spike trains in order to trigger a synaptic depression event. The evidence for this process is held in a separate accumulator (denoted Accum- in the figure). The two state machines operate independently of one another. (While an early version of this synaptic machinery used a shared accumulator with changes of opposite sign being applied by the two state machines, this led to a random walk in the accumulator with excessive number of repeats required for some synapses to commit to either a potentiation or a depression. The split accumulator scheme requires more state bits but leads to more rapid convergence).

Note that noise spikes on either the pre-synaptic or post-synaptic streams will activate one or other of the 1-bit state machines, but the presence of the matching spike to complete the pair (and triggering an update to the accumulator) is of low probability. Even if this happens the accumulator must reach its own threshold to trigger actual modification of the synaptic weight, an event whose probability we can control by setting the threshold of the accumulator appropriately. Thus, using the proposed state machines the synaptic weight is can be protected from random fluctuations due to noise processes. 

\subsubsection{Design parameters for the Synaptic State Machines} \label{DesignParamsSynapticStateMachine}
There are five parameters directly connected to the synaptic state machine:
\begin{itemize}
\item $\tau_{pot}$ - decay time constant for the pre-waiting-post state machine
\item $\tau_{dep}$ - decay time constant for the post-waiting-pre state machine
\item $T_{pot}$ - threshold for the Accum+ counter to trigger a potentiation event
\item $T_{dep}$ - threshold for thr Accum- counter to trigger a depression event
\item $Dec_{acc}$ - decay per second of accumulators towards zero (linear decrease)
\end{itemize}

For the pre-waiting-post state machine, the $\tau_{pot}$ parameter governs the spontaneous decay of state \textbf{pre-waiting-post} to \textbf{IDLE}, so that larger values lead to prolonged vigilance and tolerance to a large inter-spike interval between the pre- and the post- spikes. The related parameter $T_{pot}$ is a count of the number of pre-post- spike pairs that must be identified in a given window of time in order to trigger a potentiation event. So a larger value for this parameter would indicate a higher level of evidence is required before potentiation occurs. The parameters for the post-waiting-pre state machine function in the same fashion.

The final parameter, $Dec_{acc}$, represents the rate at which the accumulators decay towards zero, though for simplicity this is implemented as a linear decrease rather than a true decay process. Its value is not crucial but should be enough to cause a near-threshold accumulator level to reach zero (i.e. for any partially complete evidence to be forgotten) within the time required to learn a complete pattern pair. This prevents the state from one pattern learning event from interfering with many learning events in the future.

\subsection{Control of Precise Spike Time - what change to make} \label{weightChanges}

Using the synaptic state machines described in section \ref{synapticStateMachines} it is possible to identify with high probability those synapses that may contribute to carrying an association from input to output neurons. Knowing which synapses need to change leads on to the next key question: how should the synaptic weight be changed to help to elicit the required output spikes at the right times when the input pattern is presented again?

In many artificial neural networks synaptic strength is a multi-valued quantity, it can be either negative or positive and can change sign during learning. Biological synapses are known to be either excitatory or inhibitory and do not change sign during learning. However, the number of discrete values that the synaptic strength may take \emph{in vivo} is not well understood. The work here builds on a static version of associative memory called \emph{Correlation Matrix Memory} in which synapses are binary ('0' or '1'). In our formulation three defined strengths are allowed:

\begin{itemize}
\item \textbf{Baseline}: The initial strength of a synapse before any learning has taken place
\item \textbf{Potentiated}: A strengthened synapse. This was chosen to be double the baseline strength for reasons of biological plausibility
\item \textbf{Depressed}: A weakened synapse. This was chosen to be 50\% of the baseline value, again for biological plausibility
\end{itemize}

These synaptic strengths are common in many neural networks. Unique to this learning rule is the concept of a piece of state called the \emph{Lock bit}, which is initially set to zero (unlocked). Its purpose will become clear soon. First let us describe the challenges of setting and stabilising the output spike time based on some arbitrary input pattern.

Consider a group of neurons acting as an associative memory, receiving stimulation from hundreds or thousands of neurons in a lower (input) layer and learning to produce an associated output pattern. In the terminology of associative memory, the input pattern (the \emph{key}) is to be associated with the output pattern (the \emph{data}) so that when the key is next presented, this is sufficient for the memory to reproduce the associated data. An effective memory will allow many such key-data pairs to be associated and will be able to recall each and every data item given its key.

During learning the input layer is stimulated to produce the required key pattern. In our case this will be a cyclic N-of-M pattern that will repeats multiple times in succession during the learning phase.  Simultaneously, the group of output neurons will be stimulated by an external teacher that will force the output neurons to generate the required data, itself an arbitrary cyclic N-of-M pattern. Only a small fraction (say 3\%) of the neurons in this output later will need to spike for any given association. Those that do must each learn to reproduce the given output spike time whenever the same key pattern is presented and this spike time should not drift over time as the memory learns more and more associations.

Now consider a single neuron that is to fire at time $t_{fire}$ as part of an output pattern A', given input pattern A. As each iteration of pattern A plays out, the neuron will receive stimulation that passes through synapses, generating an Excitatory Post-Synaptic Current (EPSC) in the dendrite of the neuron. Initially all synapses will have baseline strength. When each current reaches the neuron it is transformed into a potential by the membrane capacitance, $C_m$. Due to the intrinsic leakage of the membrane this potential decays over time, governed by the membrane time constant, $\tau_m$ of the neuron.

Due to co-incident arrival of several such spikes in a short time window the membrane potential can rise to a fixed threshold potential, $V_{fire}$, triggering the generation of an action potential or \emph{spike} that is transmitted down the axon of the neuron to its many targets. When this occurs the membrane potential is reset to some baseline level, $v_{reset}$.

Consider an output neuron that receives a portion of the input pattern (a \emph{key}) via a set of excitatory synapses whose weights have been initialised to some baseline value. Before any learning has taken place, the stimulus provided by the input pattern will evoke changes in the membrane potential of the neuron, but if the parameters have been tuned correctly, this input stimulus will be insufficient to cause the neuron to fire.

Now imagine that this neuron has been chosen as one of the N that must be taught to fire as part of a complete \emph{data} pattern. In this case its allotted firing time is $t_{fire}$, signalled by a teaching input to the neuron at the given time (this teaching signal modelling a contextual input via the \emph{apical dendrite} in a real pyramidal neuron \cite{shepherd04}). This teaching signal is sufficient to cause the neuron to fire immediately. 

The learning process will consist of modifying some of the synaptic parameters so that the later application of the key pattern will elicit the data pattern in the absence of the teacher. For our target neuron this means producing the required spike at the time it was instructed. After the key and data patterns have been associated, the set of neurons will be expected to learn other associations and a small fraction of these will involve our target neuron.

Section \ref{synapticStateMachines} described the mechanism to identify (with high probability) all of the synapses whose inputs may carry signal spikes. These spikes will be distributed throughout the cycle time of the pattern, $T_{cycle}$. It is tempting but simplistic to blindly modify each of these synaptic weight to be in the high-valued \textbf{Potentiated} state. This may well be sufficient to cause the output neuron to fire next time the key pattern is applied, but it does not guarantee that the spike time will be the desired time, $t_{fire}$. 

A more precise strategy is to select a sufficient subset of these identified synapses and to potentiate these, leaving the remainder at the baseline level. The selection process must be gradual to assess the impact of each new potentiated synapse before adding more. The goal is to potentiate only those synapses necessary for the neuron to fire at the desired time, $t_{fire}$ based on feed forward input alone (no teaching input). The nature of the selection process lead to synapses whose input spikes shortly precede the output spike to be identified first and those with a greater inter-spike interval with respect to the output spike are typically chosen later. (This leads to the natural emergence of the traditional STDP curves where the expected potentiation of a synapse is higher when the inter-spike interval between the pre- and post- spikes is small and lower for greater inter-spike interval, but we do not show that here).

\begin{figure}
    \centering
    \includegraphics[width=0.8\textwidth]{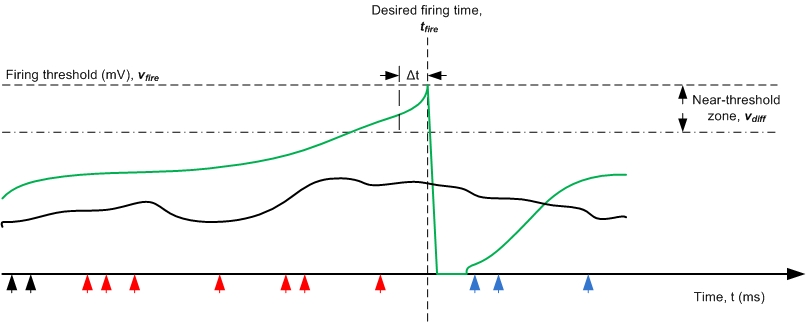}
    \caption{The evolution of the soma potential of one neuron in response to a given input pattern (arrival of spikes shown by arrow heads on the x-axis). The blue curve shows the initial response before learning has begun and the goal, through potentiation of certain synapses, is for the potential to hit the firing threshold $v_{thresh}$ of the neuron at the exact time $t_{fire}$ (green curve). The STDP-based learning rule will tend to potentiate synapses receiving input soon before the firing time (red arrows) and depress those soon after the firing time (light blue arrows).}
    \label{fig:somaPotentialEvolution}
\end{figure}

Figure \ref{fig:somaPotentialEvolution} shows the evolution of the soma potential for one output neuron, with its expected firing time $t_{fire}$ and threshold, $v_{fire}$ marked. The time of arrival of a set of identified input signal spikes is shown as red arrowheads on the x-axis. Each such tick also corresponds to a synapse that should be subject to modification. After the potentiation of the first few synapses the potential is still insufficient for the neuron to fire unaided (blue line on the graph). Recruiting more synapses, whose input spikes arrive earlier and earlier with respect to the desired output spike time, has the effect of boosting the evolving potential so that it gets closer to the threshold.

Eventually, the signal arriving at the neuron from the input alone is sufficient to fire the neuron, just on or before the required time (green trace). This is the desired end state. The next question is how to protect this carefully controlled output spike time from drift - there are other synapses that have been identified as carrying input pattern that have not yet been potentiated. If any of these are left free to be selected and potentiated by another pattern association at a later time this will affect the output spike time when the current key pattern is subsequently re-applied. The solution is provided by a conceptually new piece of synaptic state has not been anticipated in work to date on STDP, the \emph{lock bit}.

When a synapse has been selected using the Synaptic State Machines, its lock bit will be set, irrespective of whether or not its synaptic weight is potentiated. And when set, the synaptic weight can not be modified by any subsequent learning event (unless explicitly cleared, which is the process of forgetting). Thus the weight value is frozen at its given level. The effect of this is that once the set of signal carrying synapses have been identified and the weight modified as required, the response of this neuron to the same input key pattern \emph{should not change even though other associations are being stored in the same neuron}.

As noted, in order to forget an association the lock bits of the frozen synapses must be cleared (freeing up the synapse for recruitment in another association) and the synaptic weight returned to the baseline value. Notice that forgetting is \textbf{not} performed by depressing the synaptic weight, which goes against current wisdom. The depression state of the synaptic weight has a different purpose, described later in this section. Forgetting is outside the scope of the presented work since we seek to quantify the capacity of the network with no information lost.

Mechanistically, we implement this logic in the neuron code by examining the potential of the neuron on the cycle before the teaching signal triggers the neuron to fire. If the soma potential was close to threshold on this cycle (within a given tolerance band of magnitude $v_{diff}$) then we assume that the neuron would have fired without the teacher on the following cycle. In this case any synapses that are identified for potentiation or depression are locked but their synaptic weight is not changed. The performance of the network is sensitive to the value of $v_{diff}$ in our current implementation. If too low, then too many synapses do not undergo potentiation and this can cause the neuron to fire either very late or not at all when tested. If too high, too many unneeded synapses are potentiated, causing the neuron to fire too early under test. It would be desirable to reduce the sensitivity of the output firing time with respect to this parameter - this is an area of ongoing investigation.

As more and more associations are stored in a set of neurons, the background level of potential (before an input key is applied) can rise over time if only baseline and potentiated synapses are present. This creates a distortion in the output spike time if not compensated for. It is therefore desirable for the potential of the neuron to reach a value that is consistent from one iteration to the next, several milliseconds before the required output spike time so that potentiated input from this point onwards will do the work of making the neuron fire at the required time on each iteration. If this is achieved, the neuron potential should evolve with the same trajectory on each iteration (noise and jitter notwithstanding) causing the output spike time to occur at the same time within each cycle.

To achieve this goal, the learning rule will also seek out and depress some of the synapses that are identified as being carriers of the input pattern. The \textbf{post-waiting-pre} synaptic state machine has the task of locating synapses that carry input spikes arriving soon after the required firing time of the output neuron. These can be safely depressed since the input arriving through these synapses arrives too early to influence the firing of the neuron on its next cycle. Thus the potentiation applied to synapses during the learning of a new association is balanced by an equivalent amount of depression.

We will now claim that \textbf{the purpose of the depressed synapse is homeostatic}: contributing to the maintenance of a consistent background potential in the neuron soma, necessary to ensure that the output spike times do not deviate as the memory is loaded.

What level should this background potential be set to? An obvious but incorrect choice is the reset voltage, $v_{rest}$, to which the neuron potential would converge with no external input. In fact this is probably the worst choice. The rate at which the potential approaches $v_{rest}$ decreases exponentially over time so the settling time to $v_{rest}$ is long relative to other choices. Another problem with this choice is that if the neuron failed to fire on the previous iteration the potential must decay all the way from near-threshold to this resting level in less than one iteration of the pattern. A third issue with this is the potential bounce that typically occurs after the refractory period following the previous spike, due to residual EPSC that still flows into the soma even after enough has arrived to trigger firing. If $v_{rest}$ were to be the chosen background level it would be necessary to wait until this potential, too, had decayed.

A better choice is a level more than halfway from $v_{rest}$ to $v_{threshold}$. It is far from $v_{rest}$ and so the potential will fall rapidly towards it if the neuron failed to fire on the previous iteration. The post-spike bounce in potential actually aids in the process of restoring the required background level. The background level is maintained by incoming spikes (whether potentiated, depressed or baseline) that are due either to signal or to noise processes. 

Based on this work that concerns learning in artificial spiking neural networks, we propose the following hypothesis concerning the operation of real biological pyramidal neurons: we claim that \textbf{the purpose of maintaining the mean potential of a neuron near threshold is part of the solution of achieving consistent output spike times as the network learns new information.}

\subsubsection{Design parameters}
The timing component of the learning rule has its own unique parameters, as well as making use of standard neuron parameters in the PyNN language:

\begin{itemize}
    \item \textbf{$W_{baseline}$}: the initial value of the each excitatory synaptic weight
    \item \textbf{$W_{max}$}: Value of the potentiated synaptic weight (in nA).
    \item \textbf{$W_{min}$}: Value of the depressed synaptic weight (in nA).
    \item \textbf{$\tau_m$}: The decay time constant of the soma potential (in ms).
    \item \textbf{$\tau_{refrac}$}: The absolute refractory period of the neuron post firing (in ms).
    \item \textbf{$T_{cycle}$}: The cycle time of one pattern (in ms).
    \item \textbf{$D_{spike}$}: The density of spikes within each pattern, equal to N/$T_{cycle}$
    \item $N_{repeats}$: The number of times a pattern pair is presented to the network during a learning event
\end{itemize}

Inside our simulation framework, rather than specifying a depressed or potentiated synaptic weight, the requirement is to specify a fractional change with each such event as well as minimum and maximum weight value. Since we only permit a single change to the weight, the effect is the same.

The decay constant of the some potential, $\tau_m$ determines how far into the past before the neuron is intended to fire that arriving spikes may contribute to that firing, with a short time constant leading to any stored charge in the soma dissipating quickly. This defines a useful window of potentiation for the learning rule: there is no benefit of potentiating a synapse so far before the intended firing time that its effect on the soma potential has dissipated before the neuron is intended to fire.
The refractory period, $\tau_{refrac}$ is used to prevent a double spike. Once the neuron has learned to produce the desired spike based on only feedforward activity, it is possible for the teaching signal to initiate a second spike is there is any time gap between these two events. The refractory period should be long enough to ensure that in this case the neurons does not produce an action potential when the teaching event arrives.

The cycle time, $T_{cycle}$ and spike density, $D_{spike}$ are used to ensure that enough signal arrives is a short enough time window so that the neuron can reach the threshold at the required time. Longer cycle time give the neuron time to recover from one spike before the build up to the next. But if the number of neurons firing in the pattern is held constant, longer cycle time reduces the density of spikes, reducing the intensity of the signal. Since the soma is leaky, having too low a spike density prevents the neuron from ever reaching threshold even with the potentiation of every synapse that passes an EPSC during the cycle.

The value of $N_{repeats}$ controls the number of cycles of the patterns are presented. There should be sufficient repeats to give the synaptic state machines time to locate every signal-carrying synapse with high probability. A higher value also allows the thresholds $T_{pot}$ and $T_{dep}$ to be higher, increasing the vigilance of the process trying to separate signal from noise. However, more repeats means that each learning event takes longer. The time required to learn one pattern pair is equal to the product of $T_{cycle}$ and $N_{repeats}$.

\section{Experiments \& Results} \label{Experiments}
To demonstrate the proposed learning mechanism, two sets of simulations were carried out. The first closely monitors the recruitment of synapses as a single neuron learns to reproduce an output spike at a given time. The convergence of the output spike time to the required value is demonstrated, as is the evolving profile of synaptic recruitment as a function of the number of repeats of the input pattern. The second experiment applies an increasing load of pattern pairs to be stored by a single neuron and assesses the ability of that neuron to produce the required output spike as a function of the loading by recording the number of correctly recalled output spikes, for a given error tolerance.

\begin{figure}
    \centering
    \includegraphics[width=1.0\textwidth]{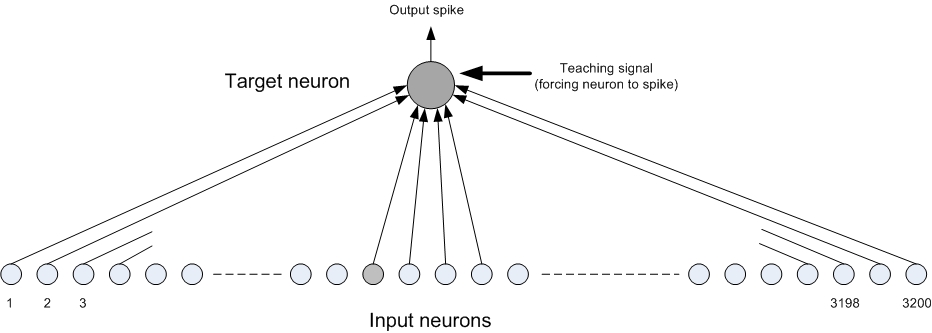}
    \caption{Network for the experiments: a single learning neuron is fed by an input layer consisting of 3,200 neurons that can deliver a cyclic N-of-M pattern. In experiment 1 the network learns to produce a spike at a given time. For experiment 2, the neuron learns to produce a list of spikes at different times in response to one of a set of input patterns.}
    \label{fig:singleNeuronNetwork}
\end{figure}

All input patterns were cyclic 75-of-3,200 patterns, of cycle length 35 ms, corresponding to an activity level of about 2.34\% in the network. The size of the population was chosen based on the observation that a mammalian pyramidal neurons has a fan-in of approximately 10,000 \cite{shepherd04}, with one third of these coming from feed-forward activity (our approximation) and the other two thirds being lateral and apical input. Thus we modeled only the feed-forward component of the fan-in. Our model also ignores any of the complexity of the dendritic tree, which is now an area of increasing research interest.

\textbf{\refstepcounter{table}\label{Tab: 01} Table \arabic{table}.}{ Network parameters for the spiking associative memory. Parameters in bold were selected to fulfil memory design requirements. Those not in bold were under-constrained and were selected to be biologically plausible.}

\begin{tabular}{llll}
\textbf{Name} & \textbf{Value} & \textbf{Description}\\
$\textbf{timestep}$ & 0.1 ms & Granularity of simulation\\
\\
$\textbf{Patterns}$ & &\\
M & 3,200 & Number of elements in each pattern\\
$P_{firing}$ & 2.34\% & Percentage of active elements in each pattern\\
N & 75 & Number of active elements in each pattern\\
$T_{cyc}$ & 35 ms & Cycle time of each pattern\\
\\
$\textbf{Populations}$ & &\\
$N_{stimulus}$ & 3,200 &  \# of input neurons carrying the stimulus pattern\\
$N_{memory}$ & 1 &  \# of output neurons carrying the response pattern\\
\\
$\textbf{Neuron and Synapse Models}$\\
$\tau_m$ & \textbf{15.0 ms} & Membrane time constant\\
$C_m$ & 30.0 nF & Membrane capacitance\\
$\tau_{ref}$ & 5.0 ms & Absolute refractory period\\
$V_{thresh}$ & \textbf{60 mV} & Firing potential\\
$V_0$ & \textbf{0 mV} & Membrane resting potential\\
$V_{reset}$ & \textbf{0 mV} & Post-action potential reset voltage\\
$E_{rev}$ & \textbf{240 mV} & Excitatory synapse reversal potential\\
$\tau_{rise}$ & 0.2 ms & EPSP rise time constant\\
$\tau_{fall}$ & 3.0 ms & EPSP fall time constant\\
$T_D$ & \textbf{1.0 ms} & Action potential back-propagation time to synapse\\
\\
$\textbf{Learning Parameters}$\\
$Dec_{acc}$ & \textbf{1.0 per sec} & Rate of decrease in magnitude of each accumulator towards zero\\

$\tau_{pot}$ & \textbf{9.6 ms} & Decay time constant of pre-waiting-post state machine\\
$\tau_{dep}$ & \textbf{11.0 ms} & Decay time constant of post-waiting-pre state machine\\
$T_{pot}$ & \textbf{+5} & Potentiation threshold\\
$T_{dep}$ & \textbf{-5} & Depression threshold\\
$V_{diff}$ & \textbf{1.0} & Lock zone for synaptic modification\\
$W_{max}$ & 0.14 pA & Maximum weight value\\
$W_{init}$ & 0.07 pA & Initial weight value\\
$W_{min}$ & 0.0 pA & Minimum weight value\\
$A_{minus}$ & \textbf{0.5} & Depression rate\\
$A_{plus}$ & \textbf{0.99} & Potentiation rate\\

\end{tabular}{}

The table lists the parameters and their values for all of the experiments. All neurons were of the conductance-based  leaky-integrate-and-fire neuron model type, developed as part of the SpiNNaker implementation of the PyNN language. The table also includes parameters for the learning rule that are unique to this work.

\subsection{Experiment I: Single Neuron Tests}
In this experiment the single neuron network (figure \ref{fig:singleNeuronNetwork}) was taught to produce a single output spike at a given time (18.0 ms into a pattern of length 35 ms). The pattern pair (input pattern and taught output spike) were presented a total of 30 times in each trial. There were one hundred independent trials.

After each iteration the current state of the synaptic weights from the 3,200 input neurons to the neuron under test were extracted from the simulation and subsequently tested to establish the output spike time that would be produced if the learning process had halted at that point. 

In each test run, the neuron under test was presented with the input pattern four times in succession and the value of the output spike time was extracted on the third iteration. This avoided the issue of the first spike which could be inaccurate, since in general the output spike could occur early in the pattern cycle before there is sufficient input to generate it accurately. Typically, the second spike is the first to be recalled reliably since the neuron will then have received a whole pattern cycle before it is required to spike. The results from all one hundred trials was averaged to produce two graphs.

\begin{figure}
    \centering
    \includegraphics[width=0.8\textwidth]{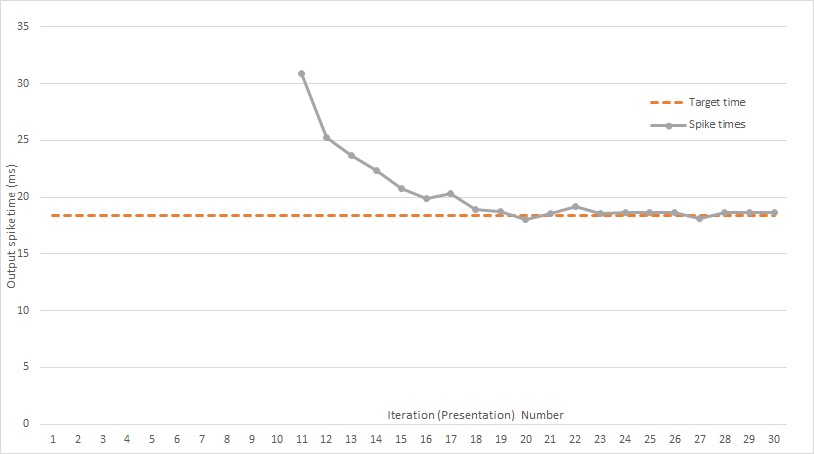}
    \caption{Graph showing the convergence of the output spike time for a single neuron, over multiple presentations of an input pattern. During the learning phase the output was stimulated at 18ms. The graph shows the output the neuron would have produced after each teaching iteration if the network were to be tested (no teacher) at that point.}
    \label{fig:convergenceNoJitterNoSpurious}
\end{figure}

Figure \ref{fig:convergenceNoJitterNoSpurious} shows the evolution of the output spike time as a function of the number of repeats of the pattern pair during training. The target output spike time is shown as an orange dotted line. For the first ten iterations no output spike is produced when tested without the teacher. The first spike that appears is very late, with subsequent iterations rapidly reducing the error. Note that once the actual and require spike times have converged, additional iterations do not lead to an increasing error, due to the change in the potentiation regime once the soma potential is in the $v_{diff}$ zone just before the required firing time.

\begin{figure}
    \centering
    \includegraphics[width=0.6\textwidth]{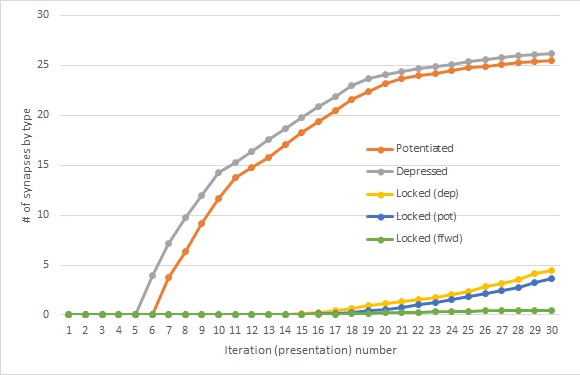}
    \caption{Graph showing the recruitment of synapses to one of five types during the learning of a single pattern association, as a function of the iteration number (the number of times the pattern pair has been presented). The original number of unrecruited synapses is 3,200. Initially, synapses are either \textbf{potentiated} or \textbf{depressed}. Later, identified synapses are locked to the baseline value when no further change in value is needed. The circumstances of the locking are differentiated: due to attempted depression (\textbf{Locked (dep)}), attempted potentiation (\textbf{Locked (pot)}) or because the neuron was already firing due to feedforward activation alone, without the aid of a teacher (\textbf{Locked(ffwd)}).}
    \label{fig:synapticProfile}
\end{figure}

Figure \ref{fig:synapticProfile} shows the time course of recruitment of the synaptic weights from their initial baseline state to one of the locked states: Potentiated, Depressed, and Locked at Baseline. We further differentiated the third category into three types, depending on the reason for remaining in the locked baseline state: Locked(dep) means that the synapse had been scheduled for depression but the soma potential was in the $v_{diff}$ region on the cycle before the teaching signal; Locked(pot) means that the synapse had been scheduled for potentiation at that time; Locked(ffwd) means that the neuron fired due to the input stimulus (feed forward activity) alone in the iteration in which the synapse was chosen for modification.

In the early iterations synapses are either fully potentiated or fully depressed, with the other synapse types only becoming relevant after convergence has been achieved. It is interesting to note that there are still full potentiation and depression events happening after convergence.  As there is no jitter in these cases, it is not caused by noise. One explanation is that the neurons is not consistently in the $v_{diff}$ zone in later cycles, perhaps due to instability in the soma potential from one cycle to the next. There may be scope to improve the reliability of the background soma potential, with better choices of the network parameters. An alternative explanation is that it is due to a choice we made to differentiate between the different locked synapse types: in order to distinguish them we gave each type a subtly different weight value, which may have had an impact on the soma potential. 

Note that the parameters were set to allow the convergence process to be clear for pedagogical purposes. It would be possible to set the parameters more aggressively to achieve a faster convergence to the correct spike output time. Too fast a convergence often led to an error in the output spike time, however - the correct selection of potentiated synapses is a form of relaxation process that needs sufficient time to converge or else it risks freezing into a sub-optimal final state. Therefore, there is a trade-off to be made between the speed of convergence and the accuracy of the final result.

\subsection{Experiment II: Single Neuron, Multiple Learnt Associations}
In this second experiment, the same single neuron model is required to learn a set of pattern associations. We created pattern sets with sizes from five to thirty patterns and for each pattern a random output spike time (between zero and 35 ms) was selected.

During training each pattern was presented thirty times while the teaching signal stimulated the neuron under test at the desired output spike time. After a pattern had been presented multiple times it was discarded until the test phase. The next pattern pair was then presented, and so on.

\begin{figure}
    \centering
    \includegraphics[width=0.7\textwidth]{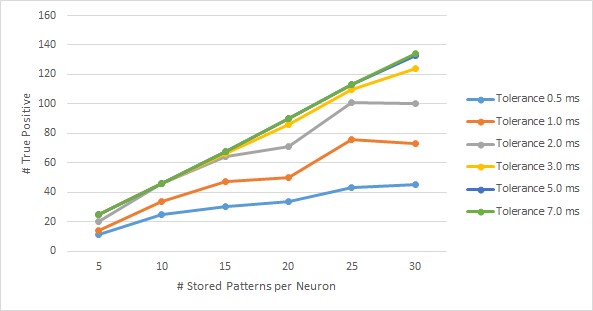}
    \caption{Graph of the number of correctly recalled output spikes for a given tolerance as a function of the number of stored pattern pairs. Patterns are stored on a single neuron. Since each pattern is presented five times, a perfect score for each size of pattern set would be five time the number of patterns.}
    \label{fig:capacityByTolerance}
\end{figure}

After a block of patterns had been memorised, the performance of the network was assessed by presenting each of the input patterns four five times and the output spike time recorded. The graph in figure \ref{fig:capacityByTolerance} illustrates the results of this recall test. There are multiple curves, each corresponding to a different tolerance to error in the output spike time. A spike is accepted as a true positive if it is within the given tolerance of the spike time provided by the teacher during training. We sought to have the majority of the spikes falling within a tolerance of 2.0 ms or less, since this implies that the spike time conveys significant temporal information in a 35 ms pattern.

For a pattern cycle time of 35 ms, a spike that meets the tolerance criteria of above 7 ms is only guaranteed to fall in the same half of the cycle as the intended spike. With this error margin, it conveys only one bit of temporal information. Despite this, it still retains information in the identity of the neurons firing, which itself conveys information but it is not a desirable outcome if the output of this memory will act as the input to the next.

We expected that the first iteration of the five during recall would give the greatest error and this was found to be the case, since the neuron has not received the full pattern at this point. Thus achieving 100\% recall was expected to be impossible.

\subsubsection{Extension to a full scale memory} The test results for a single neuron can be readily scaled up to a full scale memory consisting of 3,200 neurons and trained on entire 75-of-3,200 output patterns.

Each neuron essentially learns its required spike times independently of the others. Since the patterns are sparse (75-of-3,200 represents 2.34\% sparsity) then each neuron is only involved in firing in a small fraction of the patterns. We can therefore extrapolate to the number of patterns that can be stored and recalled in the full-scale network, which would be approximately 42 times the number of patterns that could be stored on a single neuron.

For example, from the graph, for a tolerance of +/- 3.0 ms we can store 30 patterns and expect to recall 124 of the 150 bits correctly. Most of the missing bits corresponded to the first iteration of each pattern. By accepting this proportion of missing spikes, we might expect the full scale network to store 30 x 42  = 1,260 associations between cyclic 75-of-3,200 patterns of cycle time 35 ms. This set of associations would take about 1,300 seconds (about 22 minutes) to acquire at a rate of about one per second. \textbf{We claim that this renders the associative memory a useful starting point for Short-Term Memory in the design of a larger, neural agent}, capable of storing a range of temporary memory engrams rapidly and reliably.

\section{Discussion} \label{Discussion}
We have address two major impediments to the successful use of STDP as a medium of learning in spiking neural networks:
\begin{itemize}
	\item \textbf{Stable memory using STDP} arising from the cyclic STDP learning rule
	\item \textbf{Precise spike timing} arising from the weight update procedure
\end{itemize}

In the process we have put forward a number of novel (indeed, radical) ideas on cortical functionality, based on the assumption that the precise timing of spikes is significant:
\begin{itemize}
    \item That STDP works by identifying causal links between spike streams as expressed in consistent spike \emph{pairing} from input to output.
    \item That synaptic depression is part of a mechanism to maintain precise spike timing.
    \item That the near-threshold operation of a cortical neuron is also part of a mechanism to maintain precise spike timing.
\end{itemize}

The proposed learning rule uses only locally available spike time information but leads to the stable storage of associated set of pattern pairs. We have demonstrated how a set of neurons equipped with this learning rule can be taught to reproduce spatiotemporal patterns of activity in response to previously associated key patterns, themselves spatiotemporal in nature.

A major goal of this work was to address the apparent dilemma that, while the operation of some form of STDP in biological networks is empirically accepted as fact, attempts to use this learning rule as the basis for active memory have exposed fundamental deficiencies, at least as currently formulated. By returning to the actual observed characteristics in the early STDP experiments, we have devised a learning prescription that we believe to be faithful to Nature but also potentially useful in artificial systems.

As engineers, it would be arrogant of us to claim that this work represents a true interpretation of the mechanisms at work in biological networks, but if one accepts the proviso that the brain does place significance in the timing of individual spikes, this work sheds light on the difficulties that must be overcome to learn precise spike times in a noisy environment and to maintain this learning over a period of time that is useful to the agent. We hope that our proposed solution offers a new perspective on our collective quest to understand of the brain.

In trying to remain honest with respect to the biology, some design choices were made that would be sub-optimal if our only concern were engineering efficiency. It is perhaps worthwhile to highlight these in our own defense. While the final synaptic weight can be described using only a 2-bit value, the state machines behind it (two 1-bit state machines plus two n-bit accumulators) uses considerably more resource. Furthermore, the logic behind each state machine demands the generation of a random number for each incoming spike to determine how long the heightened state of readiness (awaiting the paired spike) will persist. The cost of random number generation is not inconsiderable. If an optimal engineering solution were sought, perhaps other choices could be made to reduce these overheads.

Furthermore the learning process as described is one-sided. Synapses are potentiated until the neuron is able to fire without a teacher, with each newly potentiated synapse edging the firing time earlier, but no explicit process delays the spike again if it becomes too early. It is possible that a second (inhibitory) neuron could work here to provide opposing input to the neuron so as to delay the output spike if it appears too early. But there are many possible ways of achieving this and it was felt that it was cleaner to present the minimum useful learning rule and leave the many extensions to further work.

\subsection{Future work}
The proposed learning rule is not a complete solution in itself, but rather a building block for meso-scale circuits of spiking neurons. There are many directions that can be taken to build on this initial platform.

\subsubsection{Implement forgetting mechanism} The lock bit is the guardian that protects each memory from degradation, but how can association be deleted? While a spontaneous decay process on the lock bit in each synapse might be sufficient, this would cause a slow failure of the neuron to spike at the required time as the support for the output spike is gradually eroded. Thus it is not ideal. Alternatively, an explicit deletion step of activating the association with the input (\emph{key}) pattern and then triggering an unlocking of all active synapses would also work (via some broadcast \emph{forget} signal). This merits further investigation.

\subsubsection{Improved control of spike time} This might be done using a second, inhibitory neuron to retard the generation of the output spike when it is generated too early.

\subsubsection{Methodology to encode arbitrary sensory information in cyclic N-of-M format} In order to interface to the real world it will be necessary to convert signals from a range of sources, including visual and auditory signals to the internal cyclic N-of-M encoding required for the learning rule. Developing this as a turnkey solution would be desirable. To facilitate, this a set of conversion rules should be devised.

\subsubsection{Building medium and long-term memory} The proposed learning rule is intended for short-term memory only. It is envisioned that parallel processes at work in every synapse could embody longer-term storage of information after filtering and possible re-coding. 

\section{Contributions}
SD developed the theory, wrote the tests and analysis code. OR contributed to the theory and parameter tuning, devised the experiments and coded the learning rule on our most recent tool chain. SBF contributed early discussions on the learning process.

\section{Acknowledgements}
Funding: SD and SBF were funded by the European Union under grant no. ERC-320689 (BIMPC). The design and construction of the SpiNNaker machine was supported by EPSRC grant EP/G015740/1 (BIMPA) with ongoing platform software development supported by the European Union under grant no. FP7-604102 (HBP).

Several colleagues and students implemented versions of the learning rule on different incarnations of our SpiNNaker tool chain, including Andrew Rowley, Jamie Knight and Sergio Davies. Michael Hopkins implemented an early version of the synaptic state machine in Mathematica and contributed to early discussions. We acknowledge their contributions to these early evaluations.

\bibliographystyle{abbrv}
\bibliography{stdp_recurrent_paper_v6_0}
\end{document}